\documentclass[]{interact}

\usepackage{adjustbox}
\usepackage[ruled,vlined]{algorithm2e}
\usepackage{amsmath}
\usepackage{bm}
\usepackage{color}
\usepackage{comment}
\usepackage{multirow}
\usepackage[round]{natbib}
\usepackage{soul}
\usepackage{tabu}
\usepackage{url}
\usepackage{xcolor}

\DeclareFontFamily{OT1}{pzc}{}
\DeclareFontShape{OT1}{pzc}{m}{it}{<-> s * [1.10] pzcmi7t}{}
\DeclareMathAlphabet{\mathpzc}{OT1}{pzc}{m}{it}

\newcommand\norm[1]{\left\lVert#1\right\rVert}

\usepackage{epstopdf}% To incorporate .eps illustrations using PDFLaTeX, etc.
\usepackage[caption=false]{subfig}% Support for small, `sub' figures and tables
\theoremstyle{plain}% Theorem-like structures provided by amsthm.sty

\theoremstyle{definition}

\theoremstyle{remark}

\begin{document}

\title{Online Domain Adaptation for Continuous Cross-Subject Liver Viability Evaluation Based on Irregular Thermal Data}

\author{
\name{Sahand Hajifar\textsuperscript{a} and Hongyue Sun\textsuperscript{b}\thanks{CONTACT Hongyue Sun. Email: hongyues@buffalo.edu}}
\affil{\textsuperscript{a,b}Department of Industrial and Systems Engineering, University at Buffalo, Buffalo, NY 14260, USA}
}

\maketitle

\begin{abstract}
%To determine if a liver being transplanted is viable or not, we propose a real-time classification framework based on its surface thermal distribution. Due to high-dimensionality and irregularity of the thermal image data, we model them as graph signals and use graph Fourier transform coefficients as predictors in the classification. However, blindly training the classifiers and disregarding the inherent differences among livers can yield erroneous classifications. Thus, to cope with this, we present an online domain adaptation approach, which jointly learns the domain-invariant features and constructs the classifiers. Our proposed method classifies the livers accurately.

Accurate evaluation of liver viability during its procurement is a challenging issue and has traditionally been addressed by taking invasive biopsy on liver. Recently, people have started to investigate on the non-invasive evaluation of liver viability during its procurement using the liver surface thermal images. However, existing works include the background noise in the thermal images and do not consider the cross-subject heterogeneity of livers, thus the viability evaluation accuracy can be affected. In this paper, we propose to use the irregular thermal data of the pure liver region, and the cross-subject liver evaluation information (i.e., the available viability label information in cross-subject livers), for the real-time evaluation of a new liver's viability. To achieve this objective, we extract features of irregular thermal data based on tools from graph signal processing (GSP), and propose an online domain adaptation (DA) and classification framework using the GSP features of cross-subject livers. A multiconvex block coordinate descent based algorithm is designed to jointly learn the domain-invariant features during online DA and learn the classifier. Our proposed framework is applied to the liver procurement data, and classifies the liver viability accurately.

\end{abstract}


\newpage
\section{Introduction}
\label{sec: intro}
%\HS{Length of Introduction: 1.5 - 2 pages}
According to the Organ Procurement and Transplantation Network, there are over 120,000 patients needing an organ transplant as of Sep. 3rd, 2020 \citep{USdep}. Every 9 minutes, someone is added to the national transplant waiting list, and 95 transplants take place each day in the US on average \citep{USdep}. One challenge for the organ transplantation is that the maximum viability of organs under ideal procurement conditions is limited, only 4-6 hours for lungs and hearts and 8-12 hours for livers and pancreas. The actual viability of an organ is dependent on many factors such as donor health and procurement conditions. It is very important to accurately evaluate an organ's viability before its transplantation. 

In this paper, we focus on the liver disease. According to the National Center for Health Statistics, there are 4.5 million adults being diagnosed with liver disease in 2018 \citep{CDC}. An early diagnosis and treatment of liver disease can be challenging \citep{mueller2014non,badebarin2017different}, and many patients eventually suffer from end-stage liver disease. For end-stage liver disease, which accounts for 41,743 deaths in the US in 2017, liver transplantation has proven to be the most effective method to extend patients' life \citep{CDC, lan2018non}. However, after a liver being harvested from a donor, it needs to wait for a matching recipient and be transported to the recipient. During this period, the liver is usually kept in a cold box with circulating perfusion fluid for hours to reduce the decrease of functionality \citep{cameron2015organ, lan2018non}. Before transplantation, pathologists need to evaluate the liver viability and make sure it is safe to transplant the liver. 

Currently, there are mainly two approaches being used to evaluate the liver viability: visual inspection and biopsy \citep{keeffe2001liver, lan2015quantitative}. In visual inspection, the pathologists look at the surface of a liver and judge its viability. The visual inspection may suffer from the pathologists' inadequate experiences and is subject to the lack of consistent standards \citep{petrick2015utility}. In addition, the appearances of a viable and an unviable organ may not be distinguishable \citep{gao2020surface}. On the other hand, the biopsy examination and histologic analysis technique was initially used in organ transplantation in the 1980s, and has proven to be a more accurate evaluation technique \citep{vazquez2004importance}. However, the biopsy examination technique suffers from two disadvantages: (1) the judgment on biopsy results depends on the pathologists' expertise and is prone to the subjective errors; and (2) taking biopsy is an invasive procedure and causes very small but irremediable damage to the organ \citep{rothuizen2009liver,lan2015quantitative}.
It was reported that the lack of accurate non-invasive assessment methods wastes 20\% of donor organs every year \citep{lan2020organ}. 

\begin{figure}[ht!]
\centering
  \includegraphics[width=\textwidth]{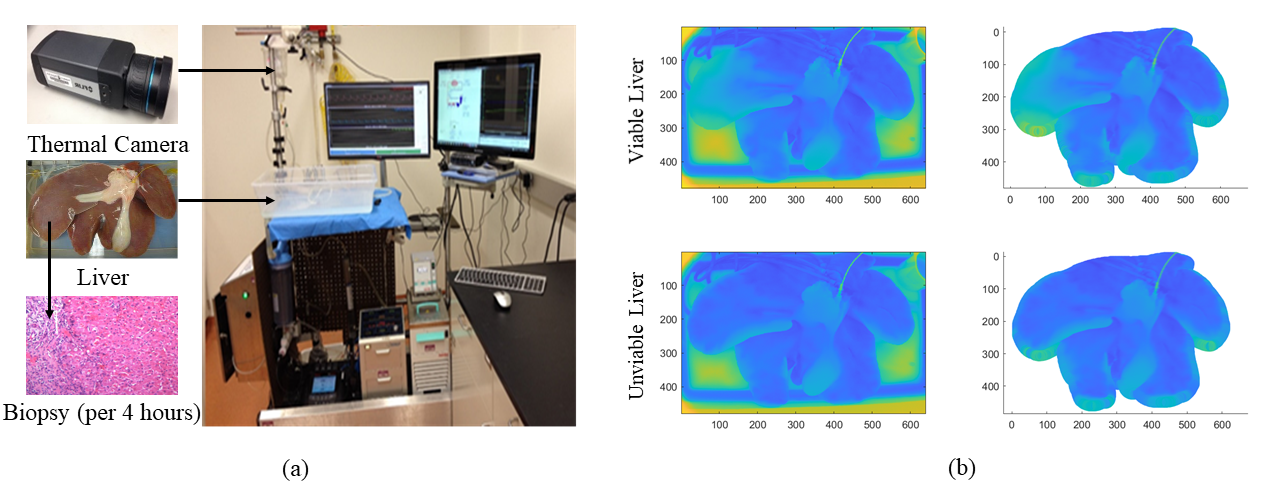}
  \caption{(a) Perfusion and non-invasive thermal imaging system \citep{lan2020organ}. Note that biopsy extraction was not a part of this work and was used for the purpose of viability evaluation; (b) Thermal images (on the left) of a viable liver versus an unviable liver, and the corresponding irregular pure liver regions (on the right)}
  \label{fig:perfusion}
\end{figure}

In recent years, infrared thermography (IRT) has gained considerable attention in medical applications as a fast, passive,
non-contact and non-invasive measuring technique \citep{lahiri2012medical}. %Fortunately, recent developments in infrared cameras and data collection and processing techniques has paved the way for online infrared image processing \citep{lahiri2012medical}.
For instance, a research team from Virginia Tech designed a machine perfusion and corresponding non-invasive thermal imaging system to measure the surface temperature of livers, as shown in Figure \ref{fig:perfusion} (a). Several analyses were performed for the liver viability assessment, and revealed a close relationship between the organ surface temperatures and viability based on this system \citep{lan2015quantitative, lan2018non, gao2018variance, gao2020surface}. As will be discussed in detail in the literature review, existing works on thermal imaging based organ viability evaluation have the limitations of: (1) using the full thermal images and thus including the background data in addition to the pure liver region (shown in Figure \ref{fig:perfusion}(b)) in the analysis. The background data may vary over time, and are irrelevant to the liver viability evaluation; (2) only considering the data in one liver or directly combining data from multiple livers in the analysis without considering the cross-subject liver heterogeneity. The variations in donors and perfusion conditions bring in the heterogeneous thermal images among livers, which need to be explicitly considered in the analysis; (3) cannot be used for the online cross-subject liver viability evaluation. The thermal images are collected continuously over time, and a model needs to be updated to accommodate for the latest information for the liver being evaluated, which will allow for the accurate online viability evaluation.

%\HS{More to add: how to perform online, how to use existing knowledge. feature and learning in two steps}
The objective of this paper is to design an online non-invasive liver viability evaluation framework based on the pure liver region thermal data and cross-subject evaluation information. This framework leverages the thermal data and biopsy evaluation information from heterogeneous livers, to predict the viability of a new liver based on its thermal images in real time without the need of taking biopsies. To achieve this objective, we face several challenges: (1) The pure liver region is irregular (irregular data format), where traditional feature and image processing methods are not directly applicable; (2) The heterogeneity of livers, because of different donors and machine perfusion conditions, prevent us from directly training a model for cross-subject liver viability evaluation; and (3) The continuously collected thermal images contain rich information on the liver condition, and one needs to update the model with the streaming images while still achieve the online viability evaluation. To address these challenges, we propose a novel online domain adaptation (DA) and classification framework based on the graph signal processing (GSP) features of irregular pure liver thermal data. The contributions of the proposed framework are: 
\begin{itemize}
    \item We propose to model the irregular thermal data of the pure liver region by graph signal processing for the first time, and use graph Fourier transform (GFT) to extract the graph features from the irregular data.
    \item We propose a novel online DA and classification framework based on the graph features for the online cross-subject liver viability evaluation. 
    \item We develop a novel multiconvex block coordinate descent based algorithm to jointly learn the domain-invariant features and the classifier.
\end{itemize}

\begin{comment}
We propose XX. The novelties of the proposed framework are XX. 
\end{comment}

The reminder of the paper is organized as follows. In Section 2, we review the relevant literature on organ viability evaluation based on thermal images, on GSP and on DA. Our proposed method is introduced in Section 3, and being demonstrated in Section 4. Finally, we conclude the paper and discuss about the future work in Section 5.

\section{Literature Review}
\label{sec: lit}
%\HS{Length of Review: around 2 pages}
\subsection{Organ Viability Evaluation based on Thermal Images}

IRT has been applied to a variety of applications including, but not limited to, kidney transplantation, brain imaging, breast cancer and liver disease \citep{lahiri2012medical}. Researchers have revealed a close association between the histomorphologic quality and surface temperatures of the organ \citep{gorbach2009assessment,skowno2014near,vidal2014near,lan2015quantitative,kochan2015ft, lan2018non, gao2018variance,gao2020surface}. 

%For instance, \cite{gorbach2008assessment} used infrared imaging to explain renal ischemia based on renal surface temperature and discovered that the blood flow oscillations measured through infrared imaging can be used as an early indicator of renal ischemia. 
Based on the collected data, different statistical methods can be utilized for organ viability evaluation. Cox regression models were applied to determine the main factors that would result into liver failure in the post-transplant period \citep{feng2006characteristics}. A hierarchical regression method was utilized to estimate donor risk index based on patient and transplant center characteristics \citep{volk2011variation}. \cite{lan2015quantitative} applied quantitative and qualitative models to predict the number of dead cells in a liver and viability of the liver. \cite{lan2018non} extracted the principal components from the thermal images, and performed intra liver classification with logistic regression and inter liver classification with multitask learning. The classification can distinguish the viable versus unviable images within a liver accurately but did not generalize well to multiple livers since it did not consider the cross-subject heterogeneity among livers. 
The above works did not consider the updated liver information over time thus cannot achieve online viability evaluation. \cite{gao2018variance,gao2020surface} performed the change detection of surface temperature variation for a single liver in an online manner, and found that the temperature measurements show a high variation when the liver is viable and a low variation if the liver is unviable. However, these works only detected the change for a single liver and did not borrow the valuable pathologists' evaluation information from other livers, which can be beneficial to the viability evaluation. %In addition, all of the above works used the full image and thus included the background data in addition to the pure liver thermal profile (shown in Figure 1(b)) in the analysis. The background data may vary over time, and are irrelevant for the liver viability evaluation. 

There is a lack of online organ viability evaluation method to explicitly consider the heterogeneous livers' thermal images and evaluation information, and to address the irregular thermal data of the pure liver region. 

\subsection{Graph Signal Processing}

GSP is a quickly developing subfield of signal processing that generalizes classical techniques of signal processing to irregular graph domain by considering the graph structure of multivariate signals  \citep{menoret2017evaluating}. 
%\HS{Make sure we explained what is a graph in our case in the introduction}
%GSP is developed on the notion that the eigenvectors of the graph Laplacian matrix are similar to Fourier modes; hence, it can be utilized to extract a spectral representation of signals situated on a graph by using the Graph Fourier Transform operator (GFT) \citep{menoret2017evaluating}. 
According to \cite{ortega2018graph}, GSP has seen major applications in sensor networks, image and 3-D point cloud processing, and biological networks. %, where the signals \citep{chen2014semi} or the whole graph \citep{anirudh2019bootstrapping} can be classified.

We review the related research on biological networks, where many works modeled the data as a network (i.e., graph), such as the human brain \citep{ortega2018graph,tran2019detecting}. Unlike in classical signal processing \citep{sartipi2020stockwell,kang2020multivariate}, GSP aims at learning and leveraging the underlying graph structure of the brain \citep{ortega2018graph}. In particular, the human brain activity signals can be modeled as a graph structure, where nodes and edge weights represent brain regions and structural connectivity between brain regions, respectively \citep{sporns2010networks,bullmore2012economy}. \cite{hu2016localizing} modeled the progression of brain atrophy through a diffusion model over the brain connectivity graph. \cite{hu2016matched} proposed a matched signal detection theory to be applied to the neuroimaging data classification problem of Alzheimer's disease.
%The researches in this area are intertwined with two other application areas introduced by \cite{ortega2018graph}, images processing and machine learning. For instance, motivated by GSP, \cite{hu2016matched} proposed a matched signal detection (MSD) theory to be applied to the neuroimaging data classification problem of Alzheimer's disease (AD). Similarly, \cite{hu2016localizing} modeled the progression of brain atrophy through a diffusion model over the brain connectivity graph.
%The applications of GSP in biological networks are not limited to brain networks.
\cite{pirayre2015brane,pirayre2017brane} used GSP to infer gene interactions, which play a vital role in the recognition of novel regulatory processes in cells. The application of GSP in other biological networks is not explored yet and still remains a research gap. 

In this paper, we will use the tools in GSP for the feature extraction of the irregular thermal data of pure liver region.

\subsection{Domain Adaptation}

DA, a subfield of transfer learning, can transfer the information learnt from source (training) domain to a new target (test) domain such that the costly endeavour in data recollection and model rebuilding can be removed \citep{huang2012transfer, hoffman2014continuous,khosla2012undoing,torralba2011unbiased,hoffman2014continuous,cheng2020hybrid, kontar2020minimizing}. In general, DA can be classified into semi-supervised and unsupervised DA \citep{ding2018graph}. %In semi-supervised scenario, a limited number of labeled target data is available \citep{kumar2010co, saenko2010adapting}. Although this scenario favors the DA and makes it easier, attaining even limited number of labeled target data is not always achievable in real-world problems. On the other hand, a more challenging scenario is unsupervised domain adaptation, where the target data do not have labels \citep{long2013transfer,gong2012geodesic}.
We focused on the unsupervised DA where the target data do not have labels, which is the case in our problem setting (i.e., no biopsy labels are available for the target data of a new liver). 

There are generally three categories of methods in unsupervised DA \citep{sun2019unsupervised}: (1) Minimizing some distributional discrepancy measure to yield the matching between the source and target features in a common feature space. For instance, \cite{long2015learning} minimized the maximum mean discrepancy (MMD) and \cite{long2017deep} minimized the joint MMD between source and target domains. (2) Learning generative models to transform from the source data to the target data \citep{taigman2016unsupervised,hoffman2018cycada}. 
For instance, a direct transformation was conducted on the image pixels rather than generating a common feature space \citep{sun2019unsupervised}. (3) Training a model on the labeled source data to estimate the pseudo-labels on the target data. Afterwards, another training was conducted on the most confident pseudo-labels of target data \citep{sun2019unsupervised}. This method was given different names such as self-ensembling \citep{french2017self} and co-training \citep{saito2017asymmetric}.

The above methods cannot achieve online DA. Online DA has gained increasing attention in recent years, where the target data is provided in a streaming manner \citep{hoffman2014continuous,bitarafan2016incremental}. \cite{hoffman2014continuous} posed the question of ``what happens when test
data not only differs from training data, but differs from it
in a continually evolving way?". The authors proposed continuous domain adaptation to address the scenario of evolving domains in a classification setting. However, their method first learned domain-invariant features
and then trained classifiers in two steps, which may yield sub-optimal classification accuracy.

In this paper, we propose an online DA and classification framework that can jointly learn the domain-invariant features
and the classifier.

\section{Proposed Framework}
\label{sec: prop}
\begin{figure}[ht!]
\centering
  \includegraphics[width=\textwidth]{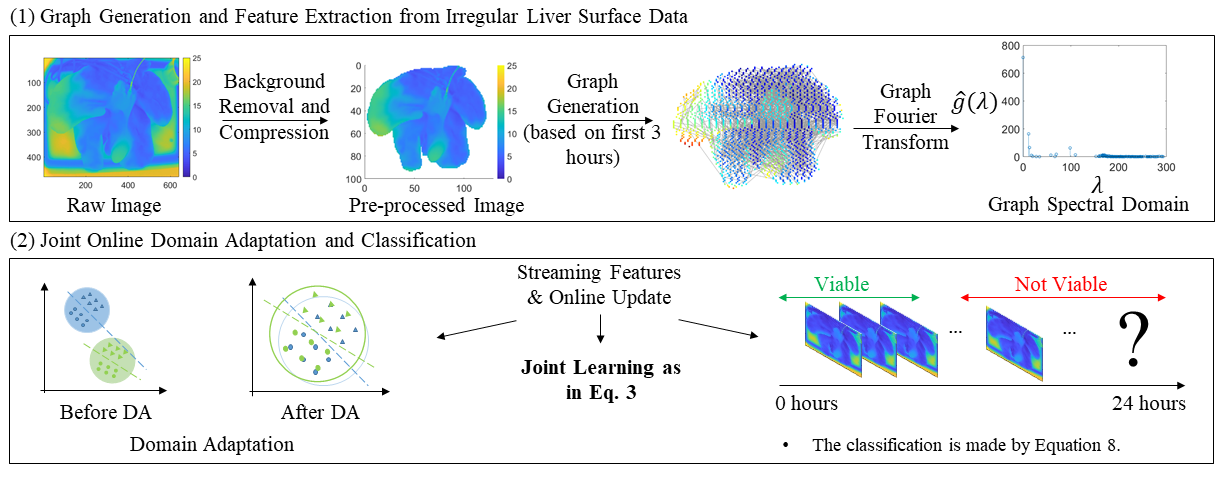}
  \caption{An illustration of the proposed framework}
  \label{fig:overview}
\end{figure}

An illustration of the proposed framework is shown in Figure \ref{fig:overview}, which includes (1) graph generation and feature extraction from irregular liver surface thermal data, and (2) joint online domain adaptation and classification. Firstly, we remove the background and extract the pure liver region from the raw liver thermal images, and then downsample the images to reduce the dimension. Thus, the later analysis is free from the background noise but is challenged by the irregular pure liver region. To address this challenge, we construct the graph for GSP of the irregular thermal data, and then extract the features from the irregular data using GFT. Secondly, we propose a novel method to jointly perform the online DA and classification for the cross-subject liver viability classification with continuous model update. In the following, we will first introduce the GSP for irregular data analysis in Section 3.1, then describe our graph construction in Section 3.2, and finally introduce the proposed online DA and classification and its joint optimization algorithm in Sections 3.3 and 3.4, respectively.

\subsection{Graph Signal Processing}
\begin{comment}
GSP extracts the spectrum information from the irregular data. The basic principle is XX, and the properties are XX.
\end{comment}

Throughout this paper, let $\mathcal{G}$ be a weighted graph containing a set $\mathcal{V}$ of $M$ vertices ($\mathcal{V} = \{ \mathpzc{v}_1,\ldots,\mathpzc{v}_M\}$). In our application, the vertices correspond to pixels of interest in the liver thermal data (i.e., the pixels after background removal and image compression). Let $\bm{A}$ be the symmetric adjacency matrix of the graph, where $A_{ij} \in \mathbb{R}^+$ represents the weight between vertices $\mathpzc{v}_i$ and $\mathpzc{v}_j$. The Laplacian matrix of graph $\mathcal{G}$ is denoted by $\bm{L}$ and calculated by $\bm{L} = \bm{D}-\bm{A}$, where $\bm{D}$ denotes the diagonal matrix of degrees obtained by $\forall i: D_{ii} = \sum_j A_{ij}$. $\bm{L}$ is symmetric and real-valued and can be factorized as $\bm{L} = \bm{F}\bm{\Lambda}\bm{F}^T$, such that $\bm{F}$ is an orthonormal matrix, and $\bm{\Lambda}$ is a diagonal matrix, whose  diagonal consists of the eigenvalues of $\bm{L}$ in ascending order. 

This factorization of the Laplacian matrix serves as the building block of GFT, which extracts a spectral representation of signals situated on a graph \citep{menoret2017evaluating}. In particular, let $\bm{s}_t\in \mathbb{R}^M$ be a vector, which denotes a realization of signal at time $t$ over the $M$ vertices of graph $\mathcal{G}$. $\bm{s}_t$ is a column of $\bm{S}$ (i.e., $\bm{S}_{.,t} = \bm{s}_t$), which is the matrix that keeps all measurements collected throughout the experiment time, where $S_{\mathpzc{v}_m,t}$ denotes the temperature of vertex $\mathpzc{v}_m$ at time $t$.

The GFT of the signal is calculated by $\hat{\bm{s}}_t = \bm{F}^T \bm{s}_t$.
The first elements of $\hat{\bm{s}}_t$, related to the lower eigenvalues in $\bm{\Lambda}$, are named low frequencies (LF) and its last elements are named high frequencies (HF). As shown in Figure \ref{fig:overview}, the HF elements are close to zero, and we consider the first $D$ LF elements to be the initial features input to DA. Actually, these LF elements preserve the main patterns in the temperature data, as shown in the case study. Let $\hat{\bm{s}}_t'$ be a vector with the first $D$ elements being the same as $\hat{\bm{s}}_t$, and zero elsewhere. We use $\hat{\bm{s}}_{t,1:D}'$ as inputs to online DA. For simplicity, we denote $\bm{f}_t = \hat{\bm{s}}_{t,1:D}'$ hereafter. 

\subsection{Graph Construction}
One needs to construct the graph $\mathcal{G}$ for GFT. 
%In general, there are three major types of graphs, geometric graphs, functional graphs and mixed graphs, used in the GSP literature. Geometric graphs model the geometric structure of a graph, whose weights are based on Gaussian kernel of the Euclidean distance between vertices without considering the pixel information \citep{menoret2017evaluating}. Functional graphs attempt to model the functional connectivity between vertices using a specific connectivity measure %, such as absolute values for correlation and covariance, and another method proposed by Kalofolias 
%\citep{kalofolias2016learn,menoret2017evaluating}. Finally, mixed graphs mix both geometric and functional connectivity of the graphs. See \cite{menoret2017evaluating} for a summary of these graphs.
% Functional graphs based on correlation and covariance, despite including the pixel information, do not allow a good reconstruction in our case. In particular, it is not possible to satisfactorily reconstruct the real signal ($\bm{s}$) by using the low frequencies (first elements in $\hat{\bm{s}}$). Moreover, Kalofolias method requires an optimization procedure, which makes its graph construction time consuming and not suitable for our case. Finally, mixed graphs which are a mixture of geometric and functional graphs inherit a combination of their characteristics. 
%proposed a new graph in which weight $A_{ij}$ models the similarity of the signals between vertices $\mathpzc{v}_i$ and $\mathpzc{v}_i$\HS{the index should be j? Please avoid such issues in your future proof-reading}. 
To preserve the main graph structure while allow the online modeling, we construct our graph as follows. Notice that GFT requires an eigendecomposition to be computed, which can be time consuming for a large matrix. We therefore construct the graph from an averaged signal $\Tilde{S}_i = \Bar{\bm{S}}_{i,1:\Tilde{t}}$ of the beginning of the data collection, so that the time-consuming eigendecompostion needs to be computed only once. Then, the weight $\bm{A}$ is defined as follows:

\begin{equation}
    \label{eq: adjacency}
    A_{i,j} = \begin{cases} \text{exp}\Big(-\frac{|\Tilde{S}_i - \Tilde{S}_j|^2}{\sigma}\Big)& i \neq j\\
    0 & i = j
    \end{cases}
\end{equation}
where $\sigma$ is a scale parameter.
This graph can provide a good reconstruction of the irregular thermal data by using the first elements of the transformed signals $\hat{\bm{s}}_t$, see details in the case study.

\begin{comment}
Data-driven methods for each time point: time consuming. 

Distance based methods: lack of spatial/pixel information.

We will use XX kernel since XX.
\end{comment}

\subsection{Online Domain Adaptation and Classification}
Let $\bm{X} \in \mathbb{R}^{N \times D}$ be the feature matrix for the graph signal $S$ of the irregular pure liver thermal data, where the $t$-th row $\bm{X}_{t,.} = \bm{f}_t^T$. $N$ is the number of samples (i.e., thermal images for a liver), and $D$ is the number of features (i.e., low frequency variables in GFT). Let $\bm{Y}$ be the class labels (viable or unviable) for thermal images. Adopting the notations from domain adaptation, we denote $\bm{X}_S$ and $\bm{X}_T$ as the source and target features, and $\bm{Y}_S$ and $\bm{Y}_T$ as the source and target labels. Our objective here is to: (1) train a classifier for the continuous target domain liver viability classification based on the labels in the source domain, and (2) find a subspace where the transformed features in source and target domains align well.

\begin{comment}
To achieve the above objectives, we design the following pieces in the objective function:

\begin{itemize}
    \item The classification loss is quantified as $L(\phi(\bm{X}_S),\bm{Y}_S|\bm{\beta})$, where $\phi(\cdot)$ is the kernel function for Reproducing Kernel Hilbert Space (RKHS) for the domain adaptation.
    \item To align the source and target domains in the subspace, we want to minimize the distance $D = MMD(\bm{X}_S,\bm{X}_T) = \norm{\frac{1}{N_S} \sum_{i=1}^{N_S}\phi(\bm{X}_{S,i}) - \frac{1}{N_T} \sum_{i=1}^{N_T}\phi(\bm{X}_{T,i})}$, where $MMD(\cdot)$ is the Maximum Mean Discrepancy (MMD) quantifying the distance of distribution in a RKHS \citep{pan2010domain,long2017deep}. The basic idea of MMD is to estimate the distance of sample distributions in source and target with distance between means of their mappings in the RKHS. After finding the mapping to minimize MMD, the features of source and target in the RKHS are expected to be close to each other.
    \item To achieve the continuous update, we learn the kernel function over time given the newly available target domain data. It is expected that the update would be smooth, and we enforce the kernel function at two close-by time to be similar, i.e., $P(t) = \norm{\bm{W}_{S,t} - \bm{W}_{S,t-1}}_F^2 = \norm{\bm{H}\bm{W}_t - \bm{H}\bm{W}_{t-1}}_F^2$, where \hl{$\bm{W}_t = \begin{bmatrix}\bm{W}_{S,t}\\\bm{W}_{T,t}\end{bmatrix}$} is the mapping matrix to be introduced later, and $\bm{H}_t = \begin{bmatrix}\bm{I}_{S\times S}&\bm{0}_{S\times t}\end{bmatrix}$. 
\end{itemize} 
\end{comment}

To achieve the objective, we propose a novel online DA and classification framework. At time $t$, we minimize the following objective function

\begin{equation}
    \label{eq: objective}
    min\: L(\phi_t(\bm{X}_S),\bm{Y}_S)+\lambda_1 D(t) + \lambda_2 P(t),
\end{equation}
where the first term $L(\phi_t(\bm{X}_S),\bm{Y}_S)$ is the classification loss at time $t$,  $\phi(\cdot)$ is the kernel function from Reproducing Kernel Hilbert Space (RKHS) for DA. The second term $D(t) = MMD(\bm{X}_S,\bm{X}_{T_t}) = \norm{\frac{1}{N_S} \sum_{i=1}^{N_S}\phi(\bm{X}_{S,i}) - \frac{1}{N_{T_t}} \sum_{i=1}^{N_{T_t}}\phi(\bm{X}_{T_t,i})}$ measures the alignment of the source domain and target domain up to time $T_t$ in the subspace $\phi(\cdot)$  \citep{pan2010domain,long2017deep}. Here, we will use the widely adopted Gaussian kernel for $\phi(\cdot)$. After finding the mapping to minimize MMD, the features of source and target domains are expected to be close to each other. Finally, the third term $P(t)$ is a penalty to enforce the function $\phi(\cdot)$ at two close-by time to be similar during the online DA. $P(t) = \norm{\bm{W}_{S_t} - \bm{W}_{S_{t-1}}}_F^2 = \norm{\bm{H}\bm{W}_t - \bm{H}\bm{W}_{t-1}}_F^2$, where $\bm{W}_t = \begin{bmatrix}\bm{W}_{S_t}\\\bm{W}_{T_t}\end{bmatrix}$ is the mapping matrix in the kernel to be defined in Section 3.4, and $\bm{H}_t = \begin{bmatrix}\bm{I}_{S\times S}&\bm{0}_{S\times t}\end{bmatrix}$. In the following, we will elaborate details of each term in Equation \ref{eq: objective}.

Firstly, to consider the potential nonlinear relationship between the features and label, we use support vector machine (SVM) as the classifier. The SVM classification loss is,
\begin{comment}
\begin{equation*}
\label{eq: svm}
\begin{gathered}
        max \: \frac{1}{\norm{\bm{\beta}}^2}\\
        subject\:to\:Y_{S,i}(\beta_0 + \phi(\bm{X}_{S,i})^T \bm{\beta}) \geq 1- \xi_i\\
        \sum_{i=1}^{N_S} \xi_i \leq C\\
        \xi_i \geq 0,
\end{gathered}
\end{equation*}
where $\xi_i$ are slack variables and $C\geq0$ is a tuning parameter. The above formulation can be transformed into the loss plus penalty format \citep{wang2008hybrid}, which solves the problem of high number of constraints in the primal problem, or variables in the dual problem \citep{lacoste2013block}. In particular, the binary SVM classification solves
\end{comment}

\begin{equation*}
\begin{gathered}
    min\:L(\phi(\bm{X}_S),\bm{Y}_S) = \:\frac{1}{N_S}\sum_{i=1}^{N_S} hinge(\phi(\bm{X}_{S,i}),\bm{Y}_{S,i}|\bm{\beta}_t,\beta_{0,t})+\frac{\lambda}{2}\norm{\bm{\beta}_t}^2 \\ =\:\frac{1}{N_S}\sum_{i=1}^{N_S}\Big(1-Y_{S,i}(\langle \bm{\beta}_t,\phi(\bm{X}_{S,i}) \rangle+\beta_{0,t})\Big)_++\frac{\lambda}{2}\norm{\bm{\beta}_t}^2
    \end{gathered}
\end{equation*}
where $\beta_{0,t}$ and $\bm{\beta}_t$ are model coefficients. According to the representer theorem \citep{scholkopf2001generalized}, $\bm{\beta}_t$ can be written as $\bm{\beta}_t = \alpha_{i,t}  \sum_{i=1}^{N_S}Y_{S,i}\phi(\bm{X}_{S,i})$. Substituting  $\bm{\beta}_t$ into the above equation, and letting $K(a,b) = \phi(a)^T \phi(b)$ and $\bm{Z}_S = diag(Y_{S,1},\ldots, Y_{S,N_S})$, we have
\begin{equation*}
\begin{gathered}
    min\:L(\phi(\bm{X}_S),\bm{Y}_S) = \:\frac{1}{N_S}\sum_{i=1}^{N_S}\Big(1-Y_{S,i}(\bm{\alpha}_t^T \bm{Z}_S K(\bm{X}_{S,i},\cdot)+\beta_{0,t})\Big)_+ +\frac{\lambda}{2}\bm{\alpha}_t^T \bm{Z}_S K(\cdot,\cdot)\bm{Z}_S \bm{\alpha}_t \\
    \:\approx \frac{1}{N_S}\sum_{i=1}^{N_S} \psi\Big(Y_{S,i}(\bm{\alpha}_t^T \bm{Z}_S K(\bm{X}_{S,i},\cdot)+\beta_{0,t})\Big) +\frac{\lambda}{2}\bm{\alpha}_t^T \bm{Z}_S K(\cdot,\cdot)\bm{Z}_S \bm{\alpha}_t
    \end{gathered}
\end{equation*}
where $\bm{\alpha}_t=(\alpha_{1,t},\ldots,\alpha_{N_s,t})^T$, and the $\cdot$ in $K(\bm{X}_{S,i},\cdot)$ and $K(\cdot,\cdot)$ refers to all the possible elements. 
Due to the nested structure of the DA transformation and SVM parameters, the above problem is bi-convex. 

We propose to develop a block-coordinate proximal gradient method based algorithm to solve the non-convex problem, which has guaranteed global convergence and high computational speed \citep{xu2013block}. However, the block-coordinate proximal gradient method requires the objective to be differentiable. We therefore use the huberized SVM, a differential approximation of the SVM formulation \citep{wang2008hybrid}, as shown in the second line of the above equation. Here, $\psi(a) = 
     \begin{cases}0&\text{for $a>1$}\\ \frac{(1-a)^2}{2\delta}&\text{for $1-\delta<a\leq1$}\\
1-a-\frac{\delta}{2}& \text{for $a\leq 1-\delta$}\end{cases}
$ is derivable, and $\delta$ is a pre-specified constant.

In summary, we have the objective function at time $t$,

\begin{equation}
    \label{eq: fullform}
    \begin{gathered}
    min\:\frac{1}{N_S}\sum_{i=1}^{N_S} \psi\Big(Y_{S,i}(\bm{\alpha}_t^T \bm{Z}_S K(\bm{X}_{S,i},\cdot)+\beta_{0,t})\Big) +\frac{\lambda}{2}\bm{\alpha}_t^T \bm{Z}_S K(\cdot,\cdot)\bm{Z}_S \bm{\alpha}_t\\
    +\lambda_1\norm{\frac{1}{N_S} \sum_{i=1}^{N_S}\phi(\bm{X}_{S,i}) - \frac{1}{N_{T_t}} \sum_{i=1}^{N_{T_t}}\phi(\bm{X}_{T_t,i})}+\lambda_2
    \norm{\bm{H}\bm{W}_t - \bm{H}\bm{W}_{t-1}}_F^2.
    \end{gathered}
\end{equation}

Next, we describe the proposed joint optimization to solve the online DA and classification in Equation \ref{eq: fullform}.

\begin{comment}
We use block proximal gradient method \citep{xu2013block} to solve the problem. The procedures (after initialization, $\bm{\alpha}$ and $\beta_0$ can start from $0$) at the $k-$th iteration are

\begin{enumerate}
    \item Solve for $(\bm{\alpha}^k,\beta_0^k )$ while fixing the other variables at values of the $(k-1)$-th iteration.
    \item 	Solve for $\bm{W}^k$, the mappings for the kernel matrix (see below for definition), while fixing the other variables at values of the $(k-1)$-th iteration. 
\end{enumerate}
\end{comment}

\subsection{Joint Optimization}
Our proposed joint optimization algorithm to solve Equation \ref{eq: fullform} at time $t$ is shown in Algorithm 1. The algorithm belongs to block-coordinate proximal descent algorithm, and will iteratively solve for $(\bm{\alpha}_t,\beta_{0,t} )$ and $\bm{W}_t$ in each iteration $k$, given the source and target inputs, source labels and initial parameters. In this section, we will introduce the details of this algorithm.

\begin{algorithm}[H]
    \SetKwInOut{Input}{Input}
    \SetKwInOut{Initialize}{Initialize}

    \Input{$\bm{X}_S,\bm{Y}_S,\bm{X}_{T_t},\lambda,\lambda_1,\lambda_2$}
    \Initialize{$\bm{u}^{-1}_t,\bm{u}^0_t$}
    \For{$k = 1,2,\ldots$}{Update $\bm{u}_t^k$ based on Eq. \ref{eq: u_update}\\
    Update $\bm{W}_t^k$ based on Eq. \ref{eq: W_update}\\
    \uIf{stopping criterion is satisfied }{
    Return($\hat{\bm{u}_t},\hat{\bm{W}}_t$) \;}
    }

    \caption{Joint Optimization Algorithm for Online DA and Classification}
\end{algorithm}

Firstly, we solve for $(\bm{\alpha}_t^k,\beta_{0,t}^k )$ in Equation \ref{eq: fullform} while fixing $\bm{W}_t^{k-1}$. In Equation \ref{eq: fullform}, only the first two terms are related to $(\bm{\alpha}_t,\beta_{0,t})$, thus to solve for $(\bm{\alpha}_t^k,\beta_{0,t}^k )$, we have

\begin{equation}
    \label{eq: u_update_0}
    \begin{gathered}
     min_{(\bm{\alpha}_t,\beta_{0,t})}\:\frac{1}{N_S}\sum_{i=1}^{N_S} \psi\Big(Y_{S,i}(\bm{\alpha}_t^T \bm{Z}_S K(\bm{X}_{S,i},\cdot)+\beta_{0,t})\Big) +\frac{\lambda}{2}\bm{\alpha}_t^T \bm{Z}_S K(\cdot,\cdot)\bm{Z}_S \bm{\alpha}_t\\
    = min_{(\bm{\alpha}_t,\beta_{0,t})}\:\frac{1}{N_S}\bm{1}^T \bm{\psi}(\bm{Z}_S \bm{VV}^T \bm{Z_S \alpha}_t+\bm{Z}_S\bm{1}\beta_{0,t})+\frac{\lambda}{2}\bm{\alpha}_t^T \bm{Z}_S \bm{VV}^T \bm{Z_S \alpha}_t \\
    = min_{\bm{u}_t}\:\frac{1}{N_S}\bm{1}^T \bm{\psi}(\Tilde{\bm{X}}_S \bm{u}_t)+\frac{\lambda}{2}\bm{u}_t^T\bm{e}^T \bm{eu}_t = min_{\bm{u}_t} f(\bm{u}_t)+\frac{\lambda}{2}\bm{u}_t^T\bm{e}^T \bm{eu}_t,
    \end{gathered}
\end{equation}

The second line is the vector form of the first line in Equation \ref{eq: u_update_0}, where $\bm{\psi}(\bm{a}) = \Big(\psi(a_1),\psi(a_2),\ldots,\psi(a_{N})\Big)^T$. In addition, the kernel matrix for the source domain $K(\bm{X}_S,\bm{X}_S)$ is approximated by $K(\bm{X}_S,\bm{X}_S) = \Phi \approx \bm{V}\bm{V}^T$, where $\bm{V \in R^{N_S\times r}}$ is a rank $r$ matrix with the $i$-th row $\bm{v}_i$ to increase the kernel computational efficiency \citep{zhang2008improved}. We further simplify Equation \ref{eq: u_update_0} as shown in its last line, where we denote $\bm{u}_t = (\beta_{0,t};\bm{\eta}_t)$, $\bm{\eta_t} = \bm{V}^T\bm{Z}_S \bm{\alpha}_t$, $\bm{e} = diag(0,\bm{1}_{r\times r})$, $\Tilde{\bm{X}}_S = (\bm{Z}_S \bm{1} \; \bm{Z}_S \bm{V})$, and $f = \frac{1}{N_S} \bm{1}^T \bm{\psi}(\hat{\bm{X}}_S \bm{u}_t)$. 

Equation \ref{eq: u_update_0} can be solved by the proximal linear update for $\bm{u}_t$ \citep{xu2013block}
\begin{equation}
    \label{eq: u_update}
    \begin{gathered}
    \bm{u}_t^k = \min_{\bm{u}_t} \langle \nabla f^k(\hat{\bm{u}}_t^{k-1}),\bm{u}_t-\hat{\bm{u}}_t^{k-1}\rangle + \frac{L_1^{k-1}}{2}\norm{\bm{u}_t-\hat{\bm{u}}_t^{k-1}}^2+\frac{\lambda}{2}\bm{u}_t^T\bm{e}^T \bm{eu}_t\\
    = \Big(L_1^{k-1} \bm{I}_{(r+1)\times (r+1)}+\lambda \bm{e}^T\bm{e}\Big)^{-1}\Big(L_1^{k-1} \hat{\bm{u}}_t^{k-1}-\nabla f^k(\hat{\bm{u}}_t^{k-1})\Big),
    \end{gathered}
\end{equation}
where $\nabla f^k(\hat{\bm{u}}_t^{k-1}) = (\Tilde{\bm{X}_S})^T \bm{\psi}'(\Tilde{\bm{X}_S} \hat{\bm{u}}_t^{k-1})$, $\hat{\bm{u}}_t^{k-1} = \bm{u}_t^{k-1}+w_1^{k-1}(\bm{u}_t^{k-1}-\bm{u}_t^{k-2})$ is an extrapolated point, $w_1^{k-1}$ is the extrapolation weight, and $L_1^{k-1}$ is the Lipschitz constant. The extrapolation can significantly accelerate the convergence speed \citep{xu2013block}.

\begin{comment}
\begin{equation}
\label{eq: u_update}
\bm{u}^k = \Big(L_1^{k-1} \bm{I}_{(r+1)\times (r+1)}+\lambda \bm{e}^T\bm{e}\Big)^{-1}\Big(L_1^{k-1} \hat{\bm{u}}^{k-1}-\nabla f^k(\hat{\bm{u}}^{k-1})\Big)
\end{equation}

has the closed form update, where $\nabla f^k(\hat{\bm{u}}^{k-1}) = \frac{1}{N_S}\bm{1}^T\bm{\psi}(\Tilde{\bm{X}_S}\bm{u}) = (\Tilde{\bm{X}_S})^T \bm{\psi}'(\Tilde{\bm{X}_S} \hat{\bm{u}}^{k-1})$.
\end{comment}

Secondly, we solve for $\bm{W}_t^{k}$, the mappings for the kernel matrix, while fixing $(\bm{\alpha}_t^k,\beta_{0,t}^k )$ in Equation \ref{eq: fullform}. Then, Equation 3 can be converted to,

\begin{comment}
min\:\frac{1}{N_S}\sum_{i=1}^{N_S} \psi\Big(Y_{S,i}[\Tilde{\bm{\alpha}}_t^k]^T \Tilde{\bm{Z}}_S\bm{K}_{\bm{X}_{S,i}}+Y_{S,i}\beta_0\Big) +\frac{\lambda}{2} [\Tilde{\bm{\alpha}}_t^k]^T \Tilde{\bm{Z}}_S\Tilde{\bm{K}} \Tilde{\bm{Z}}_S[\Tilde{\bm{\alpha}}_t^k]
    +\lambda_1 tr(\Tilde{\bm{K}}\bm{S})+\\
    \lambda_2 \norm{\bm{W}_{S,t} - \bm{W}_{S,t-1}}_F^2
    \\
\end{comment}

\begin{equation}
    \label{eq: W formulation}
    \begin{gathered}
    min_{\bm{W}_t}\:\frac{1}{N_S}\sum_{i=1}^{N_S} \psi\Big(Y_{S,i}[\Tilde{\bm{\alpha}}_t^k]^T \Tilde{\bm{Z}}_S\bm{KW}_t\bm{W}_t^T\bm{K}_{\bm{X}_{S,i}}+Y_{S,i}\beta_{0,t}^k\Big) +
    \frac{\lambda}{2} [\Tilde{\bm{\alpha}}_t^k]^T \Tilde{\bm{Z}}_S \bm{KW}_t
    \\ \bm{W}_t^T\bm{K} \Tilde{\bm{Z}}_S[\Tilde{\bm{\alpha}}_t^k]+\lambda_1 tr(\bm{KW}_t\bm{W}_t^T\bm{K}\bm{S})+\lambda_2 \norm{\bm{H}_t\bm{W}_t - \bm{H}_t\bm{W}_{t-1}}_F^2
    \end{gathered}
\end{equation}
where $\Tilde{\bm{\alpha}}_t^k = [\bm{\alpha}_t^k;\bm{0}_{N_{T_t} \times 1}]$, $\Tilde{\bm{Z}}_S = diag(\bm{Z}_S,\bm{0}_{N_{T_t} \times N_{T_t}})$, the empirical $\Tilde{\bm{K}}(\bm{X}_i,\bm{X}_j) \approx \bm{K}_{\bm{X}_i}^T\bm{W}_t\bm{W_t}^T \bm{K}_{\bm{X}_j}$, where $\bm{K}_X = [\bm{K}(\bm{X}_1,\bm{X}),\ldots,\bm{K}(\bm{X}_{S+t},\bm{X})]^T$, the empirical $\Tilde{\bm{K}} = \begin{bmatrix}\bm{K}_{S,S}&\bm{K}_{S,t}\\\bm{K}_{t,S}&\bm{K}_{t,t}\end{bmatrix} \approx \bm{KW}_t\bm{W}_t^T\bm{K}$, $\bm{S} = \begin{bmatrix}\frac{1}{N_S^2} \bm{I}_{N_S}&-\frac{1}{N_S \times N_{T_t}}\bm{I}_{N_S\times N_{T_t}}\\-\frac{1}{N_S \times N_{T_t}}\bm{I}_{N_{T_t} \times N_S}&\frac{1}{N_{T_t}^2}\bm{I}_{N_{T_t}}\end{bmatrix}$ \citep{pan2010domain}. Here, $\bm{K}_{S,S}$ is the kernel matrix of the source domain, $\bm{K}_{t,t}$ is the kernel matrix of the target domain at time $t$, and $\bm{K}_{S,t}=\bm{K}_{t,S}^T$ is the kernel matrix between the source and target domain at time $t$. The approximation to the empirical kernel can significantly reduce the number of parameters to be estimated, from the dimension of the sample to estimating $\bm{W}_t$. Then, the proximal linear update solves for

\begin{comment}
Then
\begin{equation*}
    \begin{gathered}
min\:\frac{1}{N_S}\sum_{i=1}^{N_S} \psi\Big(Y_{S,i}[\bm{\alpha}^k;\bm{0}_{N_t \times 1}]^T diag(\bm{Z}_S,\bm{0}_{N_T \times N_T})\bm{K}_{\bm{X}_{S,i}}+Y_{S,i}\beta_0\Big) \\+\frac{\lambda}{2} [\bm{\alpha}^k;\bm{0}_{N_t \times 1}]^T diag(\bm{Z}_S,\bm{0}_{N_T \times N_T})\Tilde{\bm{K}} diag(\bm{Z}_S,\bm{0}_{N_T \times N_T})[\bm{\alpha}^k;\bm{0}_{N_t \times 1}]+\\\lambda_1 tr(\Tilde{\bm{K}}\bm{S})+\lambda_2 \norm{\bm{W}_{S,t} - \bm{W}_{S,t-1}}_F^2
\\= min\:\frac{1}{N_S}\sum_{i=1}^{N_S} \psi\Big(Y_{S,i}[\bm{\alpha}^k;\bm{0}_{N_t \times 1}]^T diag(\bm{Z}_S,\bm{0}_{N_T \times N_T})\bm{KW}_t\bm{W}_t^T\bm{K}_{\bm{X}_{S,i}}+Y_{S,i}\beta_0\Big) \\+\frac{\lambda}{2} [\bm{\alpha}^k;\bm{0}_{N_t \times 1}]^T diag(\bm{Z}_S,\bm{0}_{N_T \times N_T}) \bm{KW}_t\bm{W}_t^T\bm{K} diag(\bm{Z}_S,\bm{0}_{N_T \times N_T})[\bm{\alpha}^k;\bm{0}_{N_t \times 1}]+\\\lambda_1 tr(\bm{KW}_t\bm{W}_t^T\bm{K}\bm{S})+\lambda_2 \norm{\bm{H}_t\bm{W}_t - \bm{H}_t\bm{W}_{t-1}}_F^2
    \end{gathered}
\end{equation*}
\end{comment}

$$\bm{W}_t^k = \min_{\bm{W}_t} \langle \nabla g^k(\hat{\bm{W}}_t^{k-1}),\bm{W}_t - \hat{\bm{W}}_t^{k-1}\rangle+\frac{L_2^{k-1}}{2}\norm{\bm{W}_t - \hat{\bm{W}}_t^{k-1}}^2+
\lambda_2 \norm{\bm{H}_t\bm{W}_t - \bm{H}_{t-1}\bm{W}_{t-1}}_F^2$$
where $g$ is the first three items of Equation \ref{eq: W formulation}, $\hat{\bm{W}}_t^{k-1} = \bm{W}_t^{k-1}+w_2^{k-1}(\bm{W}_t^{k-1}-\bm{W}_t^{k-2})$ is an extrapolated point, $w_2^{k-1}$ is the extrapolation weight and $L_2^{k-1}$ is Lipschitz constant.

After taking derivative of the above equation with regard to $\bm{W}_t$ and setting it to zero, we have the update for $\bm{W}_t$ as follows, see the Appendix for details.

\begin{equation}
    \label{eq: W_update}
    \begin{gathered}
    \bm{W}_t^k = (L_2^{k-1}\bm{I}+2\lambda_2 \bm{H}_t^T \bm{H}_t)^{-1}\bigg(L_2^{k-1}\hat{\bm{W}}_t^{k-1}+2\lambda_2 \bm{H}_t^T \bm{H}_{t-1}\bm{W}_{t-1} 
    -\\
    \frac{1}{N_S}\sum_{i=1}^{N_S} \psi'\Big(Y_{S,i}
    [\Tilde{\bm{\alpha}}_t^k]^T \Tilde{\bm{Z}}_S\bm{K}\hat{\bm{W}}_t^{k-1}(\hat{\bm{W}}_t^{k-1})^T\bm{K}_{\bm{X}_{S,i}}+Y_{S,i}\beta_{0,t}^k\Big) \\
    \Big(\bm{K}_{\bm{X}_{S,i}} Y_{S,i}[\Tilde{\bm{\alpha}}_t^k]^T \Tilde{\bm{Z}}_S\bm{K} +  \bm{K}\Tilde{\bm{Z}}_S[\Tilde{\bm{\alpha}}_t^k]
    Y_{S,i}
    \bm{K}_{\bm{X}_{S,i}}^T\Big) 
    \hat{\bm{W}}_t^{k-1}
    -  \lambda(\bm{K}\Tilde{\bm{Z}}_S[\Tilde{\bm{\alpha}}_t^k])\\
    (\bm{K}\Tilde{\bm{Z}}_S[\Tilde{\bm{\alpha}}_t^k])^T \hat{\bm{W}}_t^{k-1}
    -2\lambda_1 \bm{K}\bm{S}\bm{K}\hat{\bm{W}}_t^{k-1}\bigg)
    \end{gathered}
\end{equation}

Finally, after the algorithm converges, the classification at time $t$ for the new liver viability evaluation is made by
\begin{equation}
sign\big([\hat{\bm{\alpha}}_t;\bm{0}_{N_t \times 1}]^T \Tilde{\bm{Z}}_S\bm{K}\hat{\bm{W}}_t\hat{\bm{W}}_t^T\bm{K}_{\bm{X}_{S,i}}+\hat{\beta}_{0,t}\big),
\end{equation}
where $\hat{\beta}_{0,t} = \hat{\bm{u}}[1]$, $\hat{\bm{\eta}} = \hat{\bm{u}}[2:end]$ and $\hat{\bm{\alpha}}_t = (\bm{K}_{S, S} \bm{Z}_S)^{-1} \bm{V}\hat{\bm{\eta}}_t$.

\section{Case Study for Cross-subject Liver Viability Classification}
\label{sec: case}
\subsection{Porcine Liver Procurement Data}
\label{sec: data_collection}

We applied our proposed online DA and classification framework to the cross-liver 
viability classification in experimental data for porcine liver procurement originally reported in \cite{lan2015quantitative}. Here, we briefly introduce the experiment and data. After being harvested from the donor, the livers were kept in the perfusion system perfused with a physiologic perfusion fluid (modified Krebs' solution). In total, there were four livers in the case study, where the perfusion temperature for Liver 1 and Liver 2 was $4^{\circ}$ C, while for Liver 3 and Liver 4 was $22^{\circ}$ C. The livers were continuously monitored with an infrared camera (FLIR Systems, Boston, MA) for 24 hours to collect the surface temperature information. In particular, the resolution of the images were $640\times480$ and the sampling frequency was 1 frame/minute. Every four hours, a biopsy was taken from the liver and judged by a pathologist on its viability. As reported in \cite{lan2015quantitative,lan2018non}, the livers usually became unviable after the 8-th hour while it was impossible to know the exact point where a liver changes from viable to unviable. In conformity with \cite{lan2015quantitative,lan2018non}, the images 3 hours before and after the 8-th hour in each liver were excluded from the classification, and a sampling frequency of 10-minute per image was used.
Interested reader is referred to \cite{lan2015quantitative,lan2018non} for more details.

\subsection{Data Pre-processing}

\begin{figure}[ht!]
\centering
  \includegraphics[width=1\textwidth]{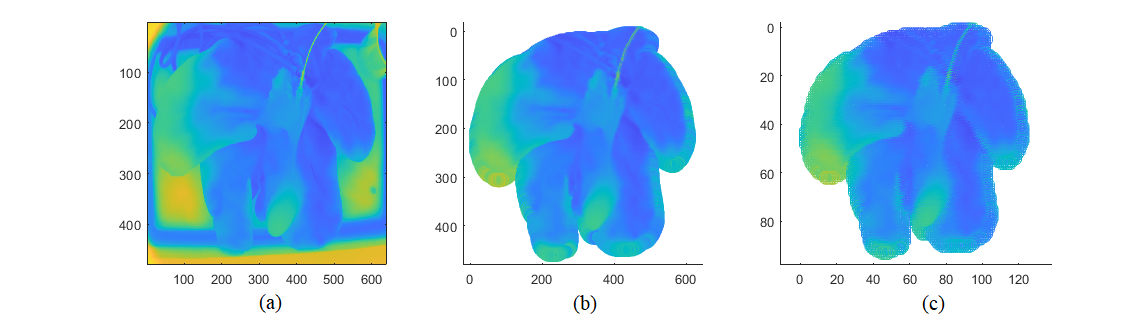}
  \caption{Data preprocessing steps: (a) Original thermal image; (b) Irregular pure liver thermal data after background removal; (c) Compressed irregular thermal data}
  \label{fig:preprocessing}
\end{figure}

The pre-processing consists of background removal and data compression of the original liver images. For a thermal image in Figure \ref{fig:preprocessing} (a), firstly, we extracted the pure liver region by defining a mask, and applying the mask to the subsequent streaming images. An irregular thermal data after the background removal is shown in Figure \ref{fig:preprocessing} (b). Afterwards, we performed the data compression by taking a moving average of a $5\times5$ window. In case part of the $5\times5$ window falls outside of the liver region at a pixel, the average was performed based on those pixels of $5\times5$ window inside the liver region. As a result, Figure \ref{fig:preprocessing} (c) shows the compressed data for GSP, where each pixel was treated as a vertex in the graph (for instance, there were 7,756 vertices in Liver 1), and the temperature value over the vertices was a graph signal.

%To shed light on how the mentioned steps work, we examine these steps for Liver 1. Firstly, a manual background removal was applied to the first image of liver 1 and the mask was stored. Applying the mask to the streaming images, resulted into $640\times480$ matrices including $188,905$ temperature values and $118,295$ NaN values. Secondly, image compression was utilized to give $128\times96$ matrices including $7,756$ temperature values and $4,532$ NaN values. Finally, the $7,756$ temperature values were vectorized to give the signals over graph. Although signal $\bm{s}$ changes over time, its dimension remains constant.

\subsection{Irregular Liver Thermal Data Feature Extraction using Graph Signal Processing}
We introduce the graph construction and GFT to extract the features from the irregular thermal data in this section. Firstly, to construct the graph, the images measured over the first three hours of the experiments were used to compute the averaged signal $\Tilde{\bm{S}}$ in each liver. Thereafter, Equation \ref{eq: adjacency} was used to obtain the $M\times M$ weight matrix. We then computed the Laplacian matrix and its factorization $\bm{L} = \bm{D}-\bm{A}= \bm{F}\bm{\Lambda}\bm{F}^T$. 
%(or $\hat{\bm{S}_t} = \bm{F}^T \bm{S}_{\boldsymbol{\cdot},t}$ to be more specific and index by time)
Secondly, we transformed the graph signal, using $\hat{\bm{s}}_t = \bm{F}^T \bm{s}_t$, to the graph spectral domain. The transformed signal $\hat{\bm{s}}_t$ has the same dimension as the original signal $\bm{s}_t$. However, as shown in Figure \ref{fig:overview}, only the first few elements of $\hat{\bm{s}}_t$ related to low frequencies are non-zero. In our case, the first 50 elements of $\hat{\bm{s}}_t$ were selected. This can significantly reduce the dimension of the problem in later analysis, while keep the patterns in the graph signal. 

\begin{figure}[ht!]
\centering
  \includegraphics[width=\textwidth]{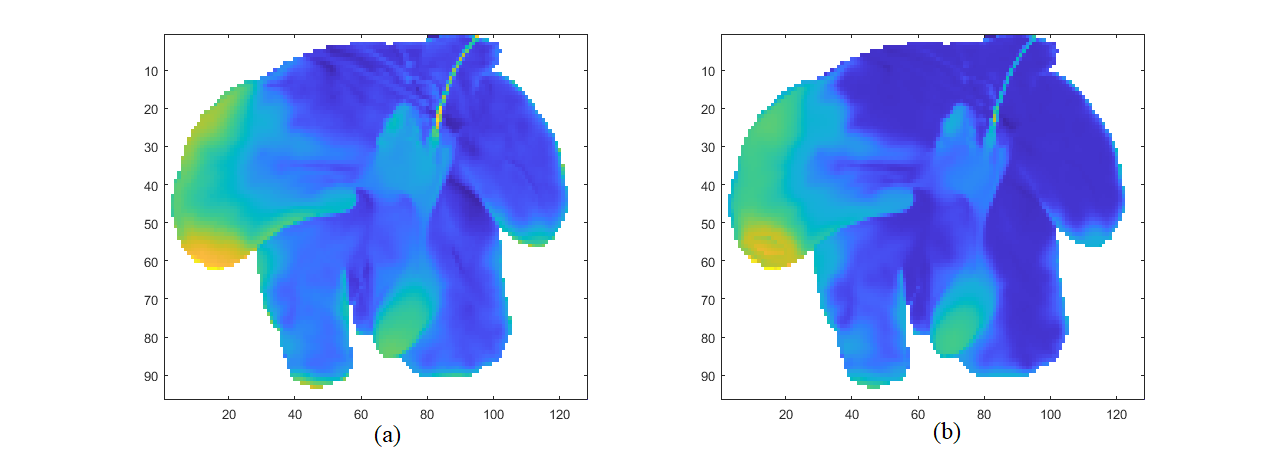}
  \caption{Illustration of irregular thermal data reconstruction of Liver 1 using: (a) All elements in $\hat{\bm{s}}_t$; (b) The first 50 elements in $\hat{\bm{s}}_t$.}
  \label{fig:recovery}
\end{figure}

Figure \ref{fig:recovery} illustrates an example of the irregular thermal data reconstruction using only the first 50 elements of $\hat{\bm{s}}_t$. The reconstruction error can be quantified by $\frac{||(\bm{F}^T)^{-1}\hat{\bm{s}}_t'-\bm{s}_t||_2}{||\bm{s}_t||_2}$, where $\hat{\bm{s}}_t'$ is a vector with the first 50 elements being the same as $\hat{\bm{s}}_t$, and zero elsewhere. On average, the reconstruction error resulted by storing first 50 elements of $\hat{\bm{s}}_t$ was 0.08 over the livers, which is reasonably small.

%Let  $\hat{\bm{s}}'$ be a vector, which has the same dimension as $\hat{\bm{s}}$, the first 50 elements of $\hat{\bm{s}}'$ are the same as $\hat{\bm{s}}$ and the remaining elements are set to zero. Using $(\bm{F}^T)^{-1}\hat{\bm{s}}'$, the signal itself can be reconstructed, where reconstruction error can be quantified by $\frac{||(\bm{F}^T)^{-1}\hat{\bm{s}}'-\bm{s}||_2}{||\bm{s}||_2}$. In our case, the average reconstruction error resulted by storing first 50 elements was \textcolor{red}{X}. Finally, the first 50 low frequency elements of $\hat{\bm{s}}$ were inputted as the initial feature of domain adaptation.

\subsection{Online Domain Adaptation and Classification via Joint Optimization}
Note that in our problem, the source domain data (both irregular thermal data and viability labels) were fully available, while the irregular thermal data in the target domain (without viability labels) were observed as a continuous data stream. We continuously made the viability classification once a streaming image is observed, and updated the model every one hour. In particular, Equation \ref{eq: fullform} was hourly optimized to update the model, and predict the labels of the forthcoming images in the target domain. The model update typically took around less than 3 seconds when performed on an HP laptop equipped with a 2.10 GHz processor and 16.0 GB of RAM running a 64-bit version of Windows 10. Here, a hourly updated was used to balance the information accumulation and domain evolution in the target domain, and other numbers can be used in other applications.

We compare our proposed method and four benchmark methods: offline SVM, online SVM, offline subspace alignment and online subspace alignment. A leave-one-out-cross-validation was used to evaluate the performance of the proposed algorithm in comparison with benchmarks, where livers were classified every 10 minutes. Specifically, in each iteration, one of the livers was considered as target and the others were considered as sources. A majority voting approach was applied, in which the final predicted label at a specific time was the one that gained more than half of the votes. 

The first two benchmarks, offline SVM and online SVM, did not benefit from any DA. In offline SVM, the underlying assumption was that all of the features associated with sources and target were known. Therefore, the features could be normalized and SVM could be applied to the normalized features. However, in online SVM, we consider the streaming nature of the target data, and the features could not be normalized and SVM was directly applied to the features obtained from GFT. For the last two benchmarks, subspace alignment (SA) is an unsupervised DA method, which was originally developed for the purpose of offline DA \citep{fernando2013unsupervised}. In this method,  the source and target features are represented by subspaces based on eigenvectors. The offline SA first perform the SA in \cite{fernando2013unsupervised}, and then perform the SVM classification. The online SA considered the streaming nature of the target data, and an SA was implemented every time that a new observation was received, followed by the SVM classification (i.e., the DA and classification are done in two separate steps).

Next, we will discuss the details of the optimization process and results comparison.

\subsubsection{Details of the Optimization Process}
\label{sec: CV}
We set $r$ to be 50 in the approximation of $K(\bm{X}_S,\bm{X}_S) $ with $\bm{V}\bm{V}^T$, and our results demonstrated that the average reconstruction error associated with this factorization was $<0.03$. In Algorithm 1, Lipschitz constants and extrapolation weights were specified as $L_1^{k-1} = 1000$, $L_2^{k-1} = 1000$, $w_1^{k-1} = 0.01$ and $w_2^{k-1} = 0.01$ \citep{xu2013block}. The $\delta$ was specified as 1 and the number of iterations was set to 50, which is sufficient for the algorithm to converge in our case study.

%In Liver 1, the source data contained 17 hours (1st hour was removed because the position of Liver 1 had changed in the 1st hour) and classification frequency was 6 image/hour. %, which translated into 102 observations (similarly, there were 108 observations considered for other livers: 6 image/hour $\times$ 18 hours = 108 images). 

We set $\lambda$ to 1 and tuned the two weights associated with DA and smoothness loss functions ($\lambda_1$ and $\lambda_2$) in Equation \ref{eq: fullform} via leave-one-out-cross-validation. In particular, we applied the leave-one-out-cross-validation to the source domain data at combinations of values of $\lambda_1$ and $\lambda_2$, and $\lambda_1$ and $\lambda_2$ were selected to maximize the average accuracy of the classification over all observation times and all left-out source domain data. Note that the tuning process only needs the source domain data, and can be performed offline. Here, the range of $\lambda_1$ and $\lambda_2$ are $\lambda_1 = [20, 40, 60, \ldots, 200]$ and $\lambda_2 = [0.002, 0.004, 0.006, \ldots, 0.020]$, respectively. The tuning parameter in SA was also selected using leave-one-out-cross-validation.

%The candidates from parameter space were specified as $\lambda_1 = [20, 40, 60, \ldots, 200]$ and $\lambda_2 = [0.002, 0.004, 0.006, \ldots, 0.020]$.

%As mentioned in Section \ref{sec: data_collection}, the images 3 hours before and after the 8th hour threshold were excluded in the classification. However, during the parameter tuning, the exclusion was not applied to the target data due to the distribution gap in the DA. %This period is only used in the learning of the domain mapping function, but will not be evaluated in the classification since their true viability conditions are not known. 

\subsubsection{Comparative Analysis}
A summary of the accuracy (ACC), precision (PRC) and recall (RCL) results for the target domain data (i.e., testing data) with different methods is given in Table \ref{tab: results}. From Table \ref{tab: results}, when there is not any DA (offline and online SVM), the accuracy ranges from 0.75 to 0.97 for offline SVM (with an average of 0.86) and from 0.35 to 0.81 for online SVM (with an average of 0.67) across the livers. On average, the recalls associated with offline and online SVM are 0.37 and 0.25, respectively. In both methods, there are cases in which NaN values are reported for precision because no livers are classified as viable in those cases. We attribute the low classification performance of these two methods to the heterogeneity of livers. However, it can be noted that normalization in offline SVM can slightly mitigate the liver-to-liver variation.  

\begin{table}[ht!]
\centering
\caption{Accuracy, precision and recall results for different methods}
\begin{adjustbox}{width=1\textwidth}
\small
\begin{tabular}{|c|ccc|ccc|ccc|ccc|ccc|}
\hline
\multirow{2}{*}{\begin{tabular}[c]{@{}c@{}}Liver\\ Number\end{tabular}} & \multicolumn{3}{c|}{\begin{tabular}[c]{@{}c@{}}SVM\\ (offline)\end{tabular}} & \multicolumn{3}{c|}{\begin{tabular}[c]{@{}c@{}}SVM\\ (online)\end{tabular}} & \multicolumn{3}{c|}{\begin{tabular}[c]{@{}c@{}}Subspace Alignment\\ (offline)\end{tabular}} & \multicolumn{3}{c|}{\begin{tabular}[c]{@{}c@{}}Subspace Alignment\\ (online)\end{tabular}} & \multicolumn{3}{c|}{\begin{tabular}[c]{@{}c@{}}Proposed\\ (our method)\end{tabular}} \\ \cline{2-16} 
                                                                        & ACC                      & PRC                     & RCL                     & ACC                     & PRC                     & RCL                     & ACC                           & PRC                          & RCL                          & ACC                          & PRC                          & RCL                          & ACC                         & PRC                        & RCL                        \\ \hline
Liver 1                                                                 & 0.81                     & NaN                    & 0.00                    & 0.81                    & NaN                    & 0.00                    & 1.00                          & 1.00                         & 1.00                         & 0.83                         & 1.00                         & 0.11                         & 0.91                        & 0.67                       & 1.00                       \\ 
Liver 2                                                                 & 0.75                     & 0.00                     & 0.00                    & 0.76                    & NaN                     & 0.00                    & 0.82                          & 1.00                         & 0.25                         & 0.82                         & 1.00                         & 0.25                         & 0.94                        & 0.80                       & 1.00                       \\ 
Liver 3                                                                 & 0.97                     & 0.96                    & 0.92                    & 0.35                    & 0.27                     & 1.00                    & 0.99                          & 1.00                         & 0.96                         & 0.76                         &NaN                         & 0.00                         & 0.79                        & 0.53                       & 1.00                       \\ 
Liver 4                                                                 & 0.89                     & 1.00                    & 0.54                    & 0.76                    & NaN                     & 0.00                    & 0.81                          & 0.56                         & 1.00                         & 0.99                         & 0.96                         & 1.00                         & 1.00                        & 1.00                       & 1.00                       \\ 
Avg.                                                                    & 0.86                     & NaN                     & 0.37                    & 0.67                    & NaN                     & 0.25                    & 0.91                          & 0.89                         & 0.80                         & 0.85                         & NaN                         & 0.34                         & 0.91                        & 0.75                       & 1.00                       \\ \hline
\end{tabular}
\end{adjustbox}
\label{tab: results}
\end{table}

When the heterogeneity is addressed by DA, the accuracy ranges from 0.79 to 1.00 (with an average of 0.91) for our algorithm, from 0.76 to 0.99 (with an average of 0.85) for online SA and from 0.81 to 1.00 (with an average of 0.91) for offline SA across the livers. Considering accuracy, precision and recall, our algorithm outperforms its online counterpart (online SA) in accuracy and recall and shows a lower precision compared to online SA (in case we ignore NaN for online SA); however, its recall is much better than that of online SA. 

If we focus on the comparison between DA methods and methods without DA, it can be noted that offline SA outperforms offline SVM, and online SA and our proposed method perform better that online SVM. This result justifies the heterogeneity of the livers and highlights the necessity of using DA to alleviate this issue. A comparison between online and offline methods shows that offline methods generally outperform online ones, which is an intuitive observation. In particular, offline SVM outperforms online SVM because it benefits from scaling through normalization. Moreover, offline SA outperforms (in two out of three measures) online SA because learning the transformation is more effective when the whole target data is available and shows a comparable performance to our proposed method. Our proposed method outperforms online SA (it shows better accuracy and significantly better recall) because of the two following reasons:

\begin{itemize}
    \item Our proposed method uses block coordinate proximal descent optimization to jointly learn the domain-invariant features and classifiers, whereas SA learns the subspaces and classifiers in a two-stage procedure.
    \item To regularize learning and improve performance, we use a loss function to assert that the mapping matrix changes slowly in time.
\end{itemize}

Finally, offline SA was the only method that showed a performance comparable to our proposed method. However, it is evident that an offline scenario is very idealistic and it is not applicable to real-world applications.

\begin{comment}
\begin{figure}[ht!]
\centering
  \includegraphics[width=3.5in]{Figures/IISE_Bar Plot.png}
  \caption{Accuracy, precision and recall results for different methods}
  \label{fig:recovery}
\end{figure}
\end{comment}

%\section{Discussions}
%\label{sec: discussions}
%\input{5_discussions}

\section{Conclusion and Future Work}
\label{sec: conclusion}
Organ transplantation is the most effective treatment for end-stage liver disease. However, the accurate evaluation of organ viability before transplantation has been a challenging issue. Traditionally, the viability is evaluated by repetitively taking invasive biopsies on the organ. Recently, viability evaluation by infrared imaging has gained increasing interest due to its noninvasive nature. However, the existing studies used the full thermal images with background noise rather than the irregular pure liver region thermal data, and did not fully address the heterogeneity of cross-subject livers during the viability evaluation.

In this paper, we propose a novel online DA and classification framework using the irregular thermal data for the liver viability evaluation. In particular, we extract features of irregular thermal data by GSP, and develop a joint optimization algorithm to jointly learn the domain-invariant features of the cross-subject livers and learn the classifiers. Our proposed method excels other benchmark methods, and has accurate online viability evaluations. 

We will pursue several directions in the future. First, in our work, in spite of having multiple sources, we considered the problem as multiple single-source DA problems and the output prediction was done based on majority voting. In the future, we will extend our framework to the multi-source DA. In addition, there have been some works on the variance change-point detection of the original thermal images \citep{gao2018variance,gao2020surface}. We will perform the graph-based change-point detection for the irregular thermal data change-point detection.

%\section*{Acknowledgments}
%\noindent The authors thank Dr. Ran Jin for sharing the data and Mr. Aditya Maunakbhai Patel for helping in data pre-processing.

\section{Appendix}
\label{sec: appendix}
We provide details on the derivation of the update of $\bm{W}_t$ in Equation \ref{eq: W_update}. 
\begin{comment}
Secondly, we solve for $\bm{W}_t^{k+1}$, the mappings for the kernel matrix, while fixing $(\bm{\alpha}_t^k,\beta_{0,t}^k )$ in Equation \ref{eq: fullform} via

\begin{equation*}
    \begin{gathered}
    min\:\frac{1}{N_S}\sum_{i=1}^{N_S} \psi\Big(Y_{S,i}[\Tilde{\bm{\alpha}}_t^k]^T \Tilde{\bm{Z}}_S\bm{K}_{\bm{X}_{S,i}}+Y_{S,i}\beta_0\Big) +\frac{\lambda}{2} [\Tilde{\bm{\alpha}}_t^k]^T \Tilde{\bm{Z}}_S\Tilde{\bm{K}} \Tilde{\bm{Z}}_S[\Tilde{\bm{\alpha}}_t^k]
    +\lambda_1 tr(\Tilde{\bm{K}}\bm{S})+\\
    \lambda_2 \norm{\bm{W}_{S,t} - \bm{W}_{S,t-1}}_F^2
    \\
     min\:\frac{1}{N_S}\sum_{i=1}^{N_S} \psi\Big(Y_{S,i}[\Tilde{\bm{\alpha}}_t^k]^T \Tilde{\bm{Z}}_S\bm{KW}_t\bm{W}_t^T\bm{K}_{\bm{X}_{S,i}}+Y_{S,i}\beta_{0,t}^k\Big) +
    \frac{\lambda}{2} [\Tilde{\bm{\alpha}}_t^k]^T \Tilde{\bm{Z}}_S \bm{KW}_t\bm{W}_t^T\bm{K} \Tilde{\bm{Z}}_S[\Tilde{\bm{\alpha}}_t^k]\\+\lambda_1 tr(\bm{KW}_t\bm{W}_t^T\bm{K}\bm{S})+\lambda_2 \norm{\bm{H}_t\bm{W}_t - \bm{H}_t\bm{W}_{t-1}}_F^2
    \end{gathered}
\end{equation*}
\end{comment}
Recall that the proximal update solves for

$$\bm{W}_t^k = \min_{\bm{W}_t} \langle \nabla g^k(\hat{\bm{W}}_t^{k-1}),\bm{W}_t - \hat{\bm{W}}_t^{k-1}\rangle+\frac{L_2^{k-1}}{2}\norm{\bm{W}_t - \hat{\bm{W}}_t^{k-1}}^2+\lambda_2 \norm{\bm{H}_t\bm{W}_t - \bm{H}_{t-1}\bm{W}_{t-1}}_F^2$$

Taking derivative of the above equation with regard to $\bm{W}_t$ and setting it to zero, we have 

\begin{equation}
    \label{eq: W_update_deri}
\nabla g^k(\hat{\bm{W}}_t^{k-1})+L_2^{k-1}(\bm{W}_t - \hat{\bm{W}}_t^{k-1})+2\lambda_2 \bm{H}_t^T(\bm{H}_t \bm{W}_t - \bm{H}_{t-1} \bm{W}_{t-1})=0, 
\end{equation}
where $\nabla g^k(\hat{\bm{W}}_t^{k-1}) = \frac{1}{N_S}\sum_{i=1}^{N_S} \frac{\partial \psi\Big(Y_{S,i}[\Tilde{\bm{\alpha}}_t^k]^T \Tilde{\bm{Z}}_S\bm{KW}_t\bm{W}_t^T\bm{K}_{\bm{X}_{S,i}}+Y_{S,i}\beta_{0,t}^k\Big)}{\partial \bm{W}_t} +\frac{\lambda}{2} \frac{\partial([\Tilde{\bm{\alpha}}_t^k ]^T \Tilde{\bm{Z}}_S \bm{KW}_t\bm{W}_t^T\bm{K} \Tilde{\bm{Z}}_S[\Tilde{\bm{\alpha}}_t^k ])}{\partial \bm{W}_t}+\lambda_1 \frac{\partial\big(tr(\bm{KW}_t\bm{W}_t^T\bm{K}\bm{S})\big)}{\partial \bm{W}_t}$. We will focus on the derivative for each of the term as follows.

According to \cite{petersen2012matrix} Eqs. 70-72, 77,

$$
\begin{gathered}
    \frac{1}{N_S}\sum_{i=1}^{N_S} \frac{\partial \psi\Big(Y_{S,i}[\Tilde{\bm{\alpha}}_t^k ]^T \Tilde{\bm{Z}}_S\bm{KW}_t\bm{W}_t^T\bm{K}_{\bm{X}_{S,i}}+Y_{S,i}\beta_{0,t}^k\Big)}{\partial \bm{W}_t}  = \frac{1}{N_S}\sum_{i=1}^{N_S} \psi'\Big(Y_{S,i}[\Tilde{\bm{\alpha}}_t^k ]^T \Tilde{\bm{Z}}_S\\
   \bm{KW}_t\bm{W}_t^T\bm{K}_{\bm{X}_{S,i}} +Y_{S,i}\beta_{0,t}^k\Big)
    \Big(\bm{K}_{\bm{X}_{S,i}} Y_{S,i}[\Tilde{\bm{\alpha}}_t^k ]^T \Tilde{\bm{Z}}_S\bm{K} + \bm{K}\Tilde{\bm{Z}}_S[\Tilde{\bm{\alpha}}_t^k ]
    Y_{S,i} \bm{K}_{\bm{X}_{S,i}}^T\Big)\bm{W}_t,
    \end{gathered}$$
    
    $$
    \begin{gathered}
    \frac{\lambda}{2} \frac{\partial\big([\Tilde{\bm{\alpha}}_t^k ]^T \Tilde{\bm{Z}}_S\bm{KW}_t\bm{W}_t^T\bm{K}\Tilde{\bm{Z}}_S[\Tilde{\bm{\alpha}}_t^k ]\big)}{\partial \bm{W}_t} = 
    \lambda(\bm{K}\Tilde{\bm{Z}}_S[\Tilde{\bm{\alpha}}_t^k ])(\bm{K}\Tilde{\bm{Z}}_S[\Tilde{\bm{\alpha}}_t^k ])^T \bm{W}_t.
\end{gathered}
$$

According to \cite{petersen2012matrix} Eq. 100,

$$\lambda_1 \frac{\partial \big(tr(\bm{KW}_t\bm{W}_t^T\bm{K}\bm{S})\big)}{\partial \bm{W}_t} = \lambda_1 \frac{\partial \big(tr(\bm{W}_t^T\bm{K}\bm{S}\bm{KW}_t)\big)}{\partial \bm{W}_t} = 2\lambda_1 \bm{K}\bm{S}\bm{KW}_t$$

Substituting the above derivative terms into Equation \ref{eq: W_update_deri}, we have

$$
\begin{gathered}
\frac{1}{N_S}\sum_{i=1}^{N_S} \psi'\Big(Y_{S,i}[\Tilde{\bm{\alpha}}_t^k ]^T \Tilde{\bm{Z}}_S\bm{K}\hat{\bm{W}}_t^{k-1}(\hat{\bm{W}}_t^{k-1})^T\bm{K}_{\bm{X}_{S,i}}+Y_{S,i}\beta_{0,t}^k\Big)
    \Big(\bm{K}_{\bm{X}_{S,i}} Y_{S,i}[\Tilde{\bm{\alpha}}_t^k ]^T \Tilde{\bm{Z}}_S\bm{K} + \\ \bm{K}\Tilde{\bm{Z}}_S[\Tilde{\bm{\alpha}}_t^k ]
    Y_{S,i} \bm{K}_{\bm{X}_{S,i}}^T\Big)\hat{\bm{W}}_t^{k-1}+
     \lambda(\bm{K}\Tilde{\bm{Z}}_S[\Tilde{\bm{\alpha}}_t^k ])(\bm{K}\Tilde{\bm{Z}}_S[\Tilde{\bm{\alpha}}_t^k ])^T \hat{\bm{W}}_t^{k-1}+
     2\lambda_1 \bm{K}\bm{S}\bm{K}\hat{\bm{W}}_t^{k-1}+ \\L_2^{k-1}(\bm{W}_t-\hat{\bm{W}}_t^{k-1})+2\lambda_2 \bm{H}_t^T (\bm{H}_t\bm{W}_t - \bm{H}_{t-1}\bm{W}_{t-1}) = 0
\end{gathered}
$$
Simplifying the equation, we have the update of $\bm{W}_t$ as shown in Equation \ref{eq: W_update}
$$
    \begin{gathered}
    \bm{W}_t = (L_2^{k-1}\bm{I}+2\lambda_2 \bm{H}_t^T \bm{H}_t)^{-1}\bigg(L_2^{k-1}\hat{\bm{W}}_t^{k-1}+2\lambda_2 \bm{H}_t^T \bm{H}_{t-1}\bm{W}_{t-1} 
    -\frac{1}{N_S}\sum_{i=1}^{N_S} \psi'\Big(Y_{S,i}\\
    [\Tilde{\bm{\alpha}}_t^k]^T \Tilde{\bm{Z}}_S\bm{K}\hat{\bm{W}}_t^{k-1}(\hat{\bm{W}}_t^{k-1})^T\bm{K}_{\bm{X}_{S,i}}+Y_{S,i}\beta_{0,t}^k\Big)
    \Big(\bm{K}_{\bm{X}_{S,i}} Y_{S,i}[\Tilde{\bm{\alpha}}_t^k]^T \Tilde{\bm{Z}}_S\bm{K} + \bm{K}\Tilde{\bm{Z}}_S[\Tilde{\bm{\alpha}}_t^k]
    Y_{S,i}\\
    \bm{K}_{\bm{X}_{S,i}}^T\Big) 
    \hat{\bm{W}}_t^{k-1}
    -  \lambda(\bm{K}\Tilde{\bm{Z}}_S[\Tilde{\bm{\alpha}}_t^k])(\bm{K}\Tilde{\bm{Z}}_S[\Tilde{\bm{\alpha}}_t^k])^T \hat{\bm{W}}_t^{k-1}
    -2\lambda_1 \bm{K}\bm{S}\bm{K}\hat{\bm{W}}_t^{k-1}\bigg).
    \end{gathered}
$$

\linespread{1} \selectfont
\label{sec:ref}
\bibliography{refs_liver.bib}
\bibliographystyle{plainnat}

\end{document}